# An Incremental Boolean Tensor Factorization approach to model Change Patterns of Objects in Images

Saritha S [1, 2, *], G Santhosh Kumar [1]

[1] *Cochin University of Science and Technology*
[2] *Rajagiri School of Engineering & Technology*
*Kochi, Kerala, India*
* Email - sarithas.sarithas@gmail.com

**ABSTRACT**

*Change detection process has recently progressed from a post-classification method to an expert knowledge interpretation process of the time-series data. The technique finds applications mainly in remote sensing images and can be utilized to analyze urbanization and monitor forest regions. In this paper, a framework to perform a knowledge based interpretation of the changes/no changes observed in a spatiotemporal domain using tensor based approaches is presented. An incremental approach to Boolean Tensor Factorization method is proposed in this work, which is adopted to model the change patterns of objects/classes as well as their associated features. The framework is evaluated under different datasets to visualize the performance for the dependency factors. The algorithm is also validated in comparison with the tradition Boolean Tensor Factorization method and the results are substantial.*

## 1. INTRODUCTION

Change detection is the process of finding changes in a time-series data [1]. The time-series data in today's world has grown immensely in terms of satellite imagery data which varies in spatial and temporal resolution. Change detection techniques applied to satellite imagery are important to analyze urbanization in environmental studies [2], forest region monitoring in earth data studies [3], damage assessment after disaster [4] and so on. Conventional approach in change detection method is to perform classification techniques on the pixel and perform an analysis on changes/ no changes [1] on the class label of the pixel. From this approach, the research has moved further away to merge homogeneous class labels and consider them as a single entity/object in the image and observe the changes happened. This technique has raised the pixel level change detection method to an object level change detection technique. Object based change detection deals with openings to evaluate different change levels by using object related characteristics like size, shape and variability. The changes happened in the object level needs to be quantified appropriately to bring out the semantic meaning of changes/no changes occurred. It might also be interesting to see the evolution of changes on a time axis to govern/assess the driving factors.

In this paper, a framework for modeling change patterns of objects in images using tensor based approach is presented. The algorithm model the data as Boolean tensors and the tensor algebraic operation of Boolean Tensor Factorization (BTF) [5] is applied to model the change patterns. The core-tensor and factor matrices obtained after Boolean Tensor Factorization helps to model the change patterns. The size of core-tensor has to be iterated repeatedly for different values to obtain very less reconstruction error. The novelty of this proposed approach is that an incremental approach to BTF is adopted to understand the change patterns of a space. In this incremental approach, the time slotted data is added incrementally to the variance of the factor matrices, than to the original factor matrices, thus extracting the changes/variances happened in each time slotted data. The major computation time to model the change patterns is attributed to solve the optimization problem of BTF and to find the variance of the factor matrices. Each factor matrix will demonstrate a weighted component of the feature associated with it, thereby giving an understanding of how an object has changed with respect to its features. The overall changes over the entire time slot is given by the core-tensor. Experiments are conducted on two sets of data and the changes occurring in the remotely sensed images are modeled. To further evaluate the algorithm in terms of the dependency factors, it is compared with the traditional BTF and the results are found to be substantial in terms of convergence time of the factorization method.

Another notable advantage associated with the incremental BTF approach as compared with the traditional BTF approach is that the storage space in the latter is 'n' tensors for 'n' time slots which is unbounded, which is saved in the former approach. However, an extra computational cost of finding variance matrices is associated with the incremental approach which can be considered negligible in comparison with the computation time of BTF.

The purpose of this paper is not to contribute about the accuracy/statistics for finding changes. The intention is to explore the modeling of changes which has resulted, so as to bridge the semantic gap in the inherent change detection process.

The reminder of the paper is organized as follows. The next section, Section 2, briefs about the relevant and related work in modeling change patterns. Section 3 describes in detail the framework proposed and the incremental BTF algorithm is presented. The evaluation of this framework is covered in Section 4. The comparison of incremental approach and traditional approach is briefed in Section 5. Section 6 provides a conclusion of this work as well as suggested future directions.

## 2. RELATED WORK

Post classification comparisons are the most wide-spread change analysis methods adopted for quantifying results of change detection. The common techniques are contingency tables and map of changed regions [6, 7, 8]. An approach to quantify change detection region wise is seen in [9], wherein the changes are modeled using contingency tables and odd ratios. In this work, a generalized Poisson regression model is used for generating a change-index to model change patterns. The visual representation of changes in the classes is presented using mosaic plots. All the above said techniques can be categorized as pixel based change detection methods. The research has moved from pixel based change detection methods to object based change detection methods, wherein the changes are sought in objects rather than pixels.

The first attempt of object based change detection (OBCD) can be seen in [10], wherein significant change between two blobs/objects are compared in gray level images using connectivity analysis. [11] is the first work which tried to incorporate geometrical and texture features into OBCD. A pioneering work in multi-temporal objects in images can be observed in [12]. In this work, the multi date image set is stored along with the related spectral features and a statistical analysis was determined to perform change detection. Similar approaches, with additional studies to characterize objects were observed in [13, 14, 15]. The change detection techniques are now more formalized to bridge the semantic gap that arise from the analysis. The temporal evolution of a land cover is supported by an external knowledge base which is equivalent to an expert interpretation in [16]. It is noted that there is very few work in the literature that supports modeling of change patterns in a semantic manner and mostly rely on post classification analysis methods. An analysis of multiple features related to an object in the context of change detection is seen in [17], in which the model of CBIR is successfully applied to the concept of change detection.

Boolean Tensor Factorization was proposed by Pauli Miettinen in 2011 [5]. The factorization approaches of tensor like CP decomposition [18] and Tucker decomposition [18] were applied for Boolean tensors also and the results were good. It is seen that the BTF is a viable method when the data is binary. There are different analysis of tensors seen in literature like offline, dynamic and stream of which can be adopted for the purpose accordingly [19].

## 3. PROPOSED DESIGN

Analysis of time series data helps to understand the change which has happened over the period of time under study. This paper attempts to analyze the change on time series image data, that is, to model change detection in images. It is assumed that the input images contain classified entities/objects. The objective is to detect changes that has happened to the image over a period of time. The workflow of the entire design is presented in Fig 1. This section is organized as described below. The first section briefs about the data taken for the study. The second subsection describes the tensor representation of the data and also explains the relevance of Boolean Tensor Factorization. The third subsection details the procedure in which the factorization method is utilized in change detection. The fourth subsection briefs the algorithm to find change detection in images using incremental approach to Boolean Tensor Factorization.

*3.1 Data Model*

The data taken for study is the time series image data. The images under study contain labelled entities/objects. The images are of a particular region/space at different time frames/slots. The different time slots attribute to the temporal features of the labelled objects in the images along with the topological information.

*3.2 Tensor Representation of Data Model*

This section explains the ways to represent the time series image data using tensors. The image dataset is considered in 'T' consecutive intervals, where each image account for each time frame. The image instant is modeled as matrix M which has 'O' rows of objects and 'F' columns of features. Features of an object can be selected depending on the domain of the image under consideration. Selection of an appropriate threshold to measure the change detection with respect to each feature associated with an object is a challenging task. Each entry in the matrix is a binary value, indicating the presence or absence of an object with the desired feature threshold values. There are 'T' matrices for the different time slots which are combined to form a 3-order tensor as $\mathcal{X} \in \mathbb{R}^{O \times F \times T}$. This can be considered as a tensor stream which is a sequence of 3-order tensor and the value of T increases with time. At every instant $t_1$, $t_2$, $t_3$,...$t_T$, a new tensor is added to the sequence. The tensor thus formed will be a Boolean type.

The design works on the principle of Boolean Tensor Factorization. The novel approach of this work lies in the fact that an incremental approach to Boolean Tensor Factorization is done to reveal the change detection of the time series data more effectively. Boolean Tensor Factorization aims to represent the tensor as a product of low order binary factors using Boolean arithmetic. This is also called tensor decomposition in tensor algebra. There are different factorization techniques of tensor, of which the important ones are CP decomposition and Tucker decomposition. In this design, the Boolean Tensor Factorization adopted is Tucker decomposition. The Boolean Tucker decomposition of the binary tensor $\mathcal{X}$ and three integers $R_1$, $R_2$, $R_3$ is given as the product of binary core tensor, $\mathcal{G}$ of size $R_1$, $R_2$, $R_3$ and binary factor matrices $A \in \mathbb{R}^{O \times R_1}, B \in \mathbb{R}^{F \times R_2}, C \in \mathbb{R}^{T \times R_3}$.

$$\mathcal{X} = \vee_{r_1=1}^{R_1} \vee_{r_2=1}^{R2} \vee_{r3=1}^{R3} \mathcal{G}(r_1, r_2, r_3).A\ B\ C$$

The factorization is optimized when

$$min\ |\ \mathcal{X} = \vee_{r_1=1}^{R_1} \vee_{r_2=1}^{R2} \vee_{r3=1}^{R3} \mathcal{G}(r_1, r_2, r_3).A\ B\ C\ |$$

Heuristic methods like ALS [20] are used to find the core tensor and factor matrices. The task is more involved than any other decomposition, due to the presence of core tensor, for whose each value can affect the binary factors. A change in a single element in $\mathcal{G}$ can completely change the product matrices. The algorithm is guaranteed to converge as the error reduces in each iteration. The core tensor and factor matrices helps to yield the underlying patterns more effectively, as the interpretation is done in low-order components.

*3.3 Interpretation of Boolean Tensor Factorization*

The Boolean factorization will yield a core tensor $\mathcal{G}$ and binary factor matrices $A \in \mathbb{R}^{O \times R_1}, B \in \mathbb{R}^{F \times R_2}, C \in \mathbb{R}^{T \times R_3}$. Each non-zero entry in the matrix $A \in \mathbb{R}^{O \times R_1}$, say $a_{ir_1}$, will indicate the presence of the i[th] object. Similarly, each non-zero entry in the matrix $B \in \mathbb{R}^{F \times R_2}$, say $b_{jr_2}$ indicates the presence of the j[th] feature associated with the objects under consideration. For the non-zero entry in the matrix $C \in \mathbb{R}^{T \times R_3}$, say $c_{kr_3}$ points to the presence of the time slot. The core tensor will indicate the presence of all these components simultaneously, that is, the non-zero entry in the core tensor will indicate the presence of an object with the threshold features at a particular time. To summarize the core-tensor will detect the presence or absence of certain objects with features (as decided per domain) at different time periods.

*3.4 Algorithm*

  This section details the algorithm for detecting changes using Boolean Tensor Factorization with incremental approach. The incremental approach introduced to Boolean Tensor Factorization contributes to the novel part of this work. The detailed steps of the algorithm are summarized as follows.

*Step 1.* The image dataset I is set on a time scale. The image dataset I consists of labeled objects at different time slots $t_1, t_2,…,t_T$. An adjacency matrix is built for all the images at different time slots $t_1, t_2,…,t_T$. The size of the adjacency matrix is O x F, where 'O' is the total number of objects in the image dataset 'I' and 'F' is the number of features associated with each object. The entry in the adjacency matrix is marked as a binary value of 1, if the particular object is present in the image with specific features exceeding the supplied threshold value, otherwise the entry is marked as 0. After this step, there will be one adjacency matrix corresponding to each time frame of the image.

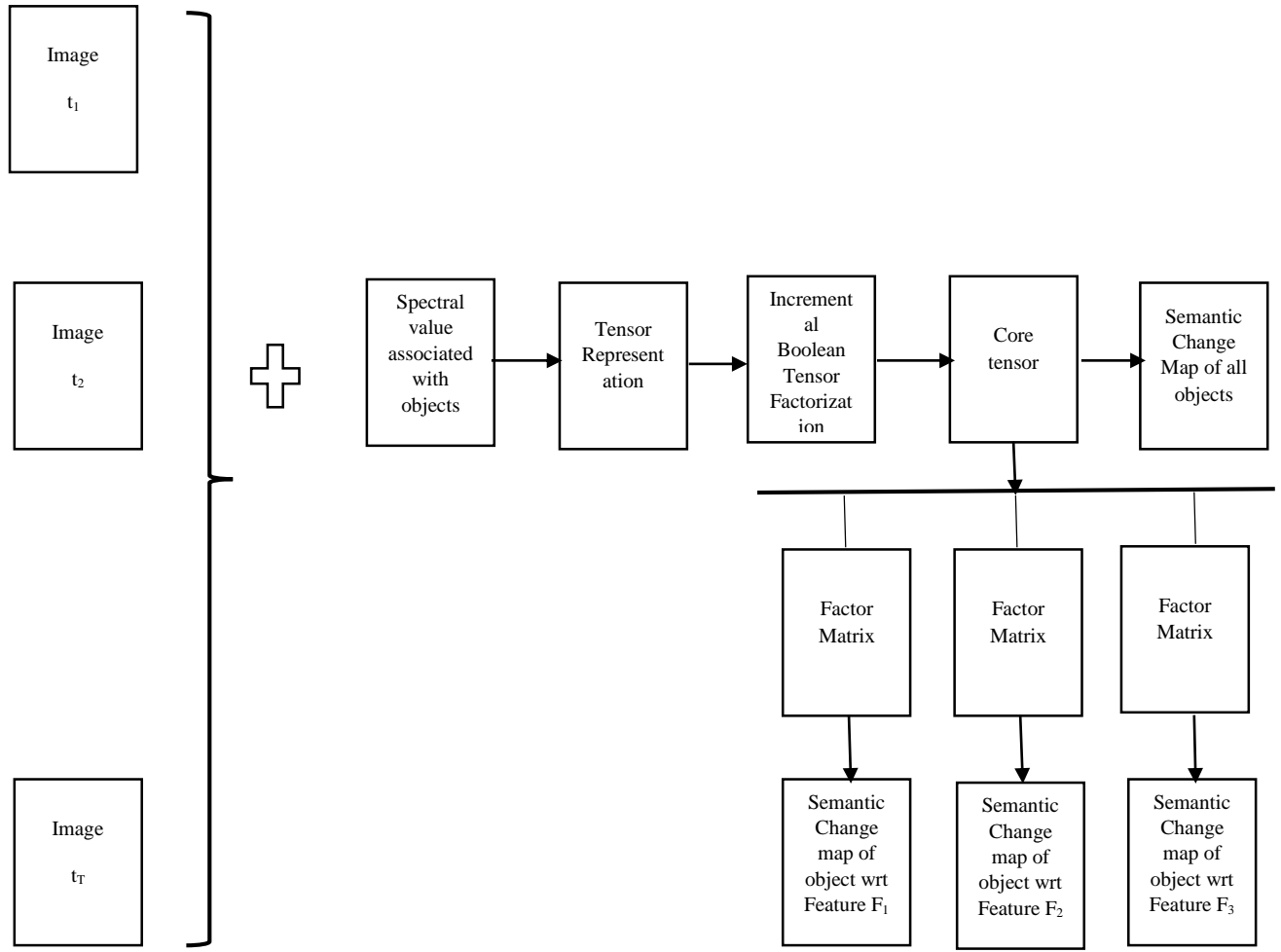

Fig 1. Workflow of the proposed design

*Step 2.* Form a 3-order tensor by combining the two adjacency matrix for time slots $t_1$ and $t_2$. This is the initial binary tensor $\mathcal{X} \in \mathbb{R}^{O \times F \times T}$.

*Step 3.* This step is the core part of the algorithm that performs incremental Boolean Tensor Factorization to obtain the core tensor. The initial steps include initialization of binary factor matrices $A \in \mathbb{R}^{O \times R_1}, B \in \mathbb{R}^{F \times R_2}, C \in \mathbb{R}^{T \times R_3}$ randomly and three integers $R_1, R_2, R_3$ which are the components for factorization and will be the dimension of the output core tensor. The initial core-tensor is all set to zeroes. For all adjacency matrices for $t = t_3, t_4, …t_T$, the factor matrices are updated as per the Boolean Tucker factorization using ALS method. After the factor matrices are updated, the covariance of each of them is calculated. The covariance matrix is updated in an incremental mode as given as follows for each matrix.

$$CA = CA_{(old)} \cdot F(T) + CA_{(new)}$$

The new covariance matrix is incrementally added to the old one. The selection of old covariance matrix is done as a function of time, $F(T)$. The function can regulate the selection of the old covariance matrices, which can be opted to be stored after each incremental addition of new time slot. The function can also be modified to take particular covariance matrices at selected time slots for finding seasonal changes which has happened in the covariance matrix. Once the covariance matrices are updated, and updation of the core tensor happens. This iteration continues until the approximation error of factorization is within the minimum error threshold value.

The advantage of the incremental Boolean Tensor Factorization lies in the fact that the space consumption is only dependent on the core tensor whose dimension is $R_1 \times R_2 \times R_3$, which will be very less than storing 'T' tensors for all the T time slots whose size will be $\mathcal{X} \in \mathbb{R}^{O \times F \times T}$ and will be highly dependent on 'T' which is unbounded. The computational cost associated with this incremental approach only lies in finding the variance matrix of the binary factors.

*Step 4.* On convergence of error value within limits, the core tensor $\mathcal{G}_{r_1, r_2, r_3}$ is produced.

**Algorithm – Incremental Boolean Tensor Factorization for change detection in images**

*Input*
    Image Dataset I
    Total number of labeled objects in the dataset O
    Total number of features associated with the objects F
    Total number of time slots in the image dataset T
    Time slot values $t_1, t_2, \ldots, t_T$
    Threshold value of features belonging to labeled objects $F_1, F_2, \ldots, F_i$
    Error Threshold $\epsilon$

*Output*
    Core Tensor $\mathcal{G}_{r_1, r_2, r_3}$

*Method*
    1. For every $I_i \in I$,
       1.1 Form an adjacency matrix M, where
           $M_{ij} = 1$, presence of object $O_i$ with feature threshold $F_j$
           $M_{ij} = 0$, otherwise
    2. Form a 3-order tensor with two adjacency matrices for time $t_1$ and $t_2$, say $\mathcal{X} \in \mathbb{R}^{O \times F \times T}$
    3. Perform incremental Boolean Tensor Factorization until the error threshold is minimum
       3.1 Randomly initialize the binary matrices $A \in \mathbb{R}^{O \times R_1}, B \in \mathbb{R}^{F \times R_2}, C \in \mathbb{R}^{T \times R_3}$, integers $R_1, R_2, R_3$
       3.2 Initialize the core-tensor $\mathcal{G}$ to all zeroes
       3.3 For $t = t_3, t_4, \ldots t_T$
          3.3.1 update_$A$ ($X$ (1), $A$, $\mathcal{G}$ (1), ( $C \odot B$ )$^T$)
          3.3.2 $CA_{(new)}$ = covariance ($A$)
          3.3.3 Update the covariance matrix, $CA = CA_{(old)} \cdot F(T) + CA_{(new)}$
          3.3.4 update_$B$ ($X$ (2), $B$, $\mathcal{G}$ (2), ( $A \odot C$ )$^T$)
          3.3.5 $CB_{(new)}$ = covariance ($B$)
          3.3.6 Update the covariance matrix, $CB = CB_{(old)} \cdot F(T) + CB_{(new)}$
          3.3.7 update_$C$ ($X$ (3), $C$, $\mathcal{G}$ (3), ( $B \odot A$ )$^T$)
          3.3.8 $CC_{(new)}$ = covariance ($C$)
          3.3.9 Update the covariance matrix, $CC = CC_{(old)} \cdot F(T) + CC_{(new)}$

3.3.10 Update_ $\mathcal{G}$ ($X$, $\mathcal{G}$, $CA$, $CB$, $CC$)

3.3.11 Calculate error as $| X - \vee_{r_1=1}^{R_1} \vee_{r_2=1}^{R2} \vee_{r3=1}^{R3} \mathcal{G}\, r_1, r_2, r_3 \,.\, CA.CB.CC |$

4. On convergence, output the core tensor $\mathcal{G}\, r_1, r_2, r_3$

## 4. RESULTS

This section describes the modeling of change patterns as per the algorithm presented in the previous section. The spatiotemporal data set used in this experiment consists of (i) Synthetic images acquired in 100 time frames consisting of 735 classes (hereafter referred as ST_Dataset_1 [21]) and (ii) Geospatial images acquired in 120 time frames (hereafter referred as ST_Dataset_2 [22]). The number of objects/classes associated with the geospatial images are a) Agricultural b) Urban c) Forest d) Road e) Bare ground (f) Beaches (g) Rivers and (h) Sea-water. The details of the dataset is summarized in Table 1.

**Table 1. Summary of Datasets used in experiments**

| Sl No | Dataset Name | No of images | No of objects/classes | Remarks |
|---|---|---|---|---|
| 1 | ST _Dataset_1 [21] | 100 | 735 | 100 time frames |
| 2 | ST _Dataset_2 [22] | 120 | 8 | 120 time frames |

For modeling changes, the proposed algorithms depends on the concept of incremental Boolean Tensor Factorization. The approximation of the original tensor is achieved through fine tuning the reconstruction error for different ranks of core tensor. Fig 2 presents the experimentation result for iterating on different values of core-tensor size to bring down the reconstruction error within the threshold value. The core tensor size of (4, 8, 8) is within the permissible value of error threshold for ST_Dataset_1. The core tensor size values of (4, 4, 4) and (4, 4, 8) are within the permissible values of error threshold for ST_Dataset_2. From the experiments, the conclusion drawn is that the core tensor size permissible within the error threshold value is highly dependent on the dataset. The intention of choosing the synthetic dataset is to prove that the incremental Boolean Tensor Factorization works theoretically correct.

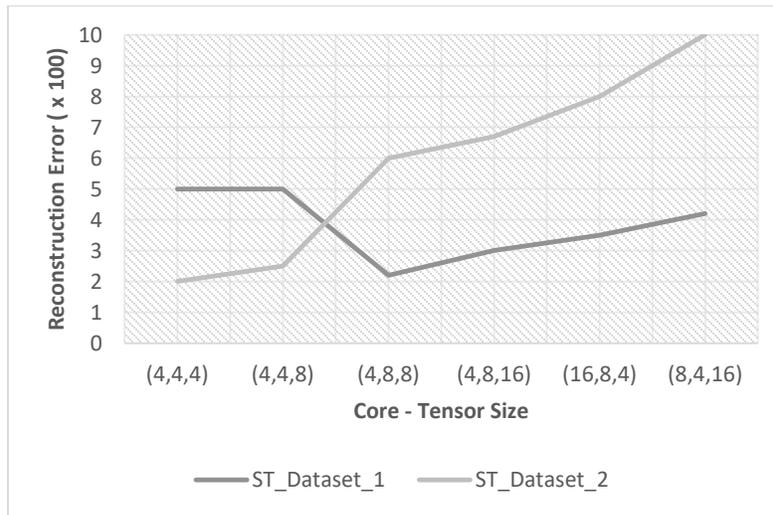

Fig 2. Optimization of core-tensor size

The change patterns are generally modeled for geospatial image datasets. Modeling of change patterns is hence forth performed only in ST_Dataset_2, as we feel that change models can be clearly understood when there are limited number of objects. The features chosen for the data set are (i) $F_1$ – length (whereas a strip length is 0 and planar length is 1) (ii) $F_2$ – shape (whereas regular is 0 and irregular is 1) and (iii) $F_3$ – texture (whereas smooth is 0 and rough is 1). Each factor matrix will be the expression of variance of each feature for the time slot under consideration. The 120 time slots are split in 10 time frames and the modeling of the feature changes for each object/class is presented in Fig 3. The variance (feature) versus time slot will model the change of features which has happened on that particular class. The variance value (between 0 to 1) will indicate the from-change or to-change of the particular class with respect to the feature (length, shape, texture). Thus measuring the feature in this aspect will give a more insight into the change patterns modeled.

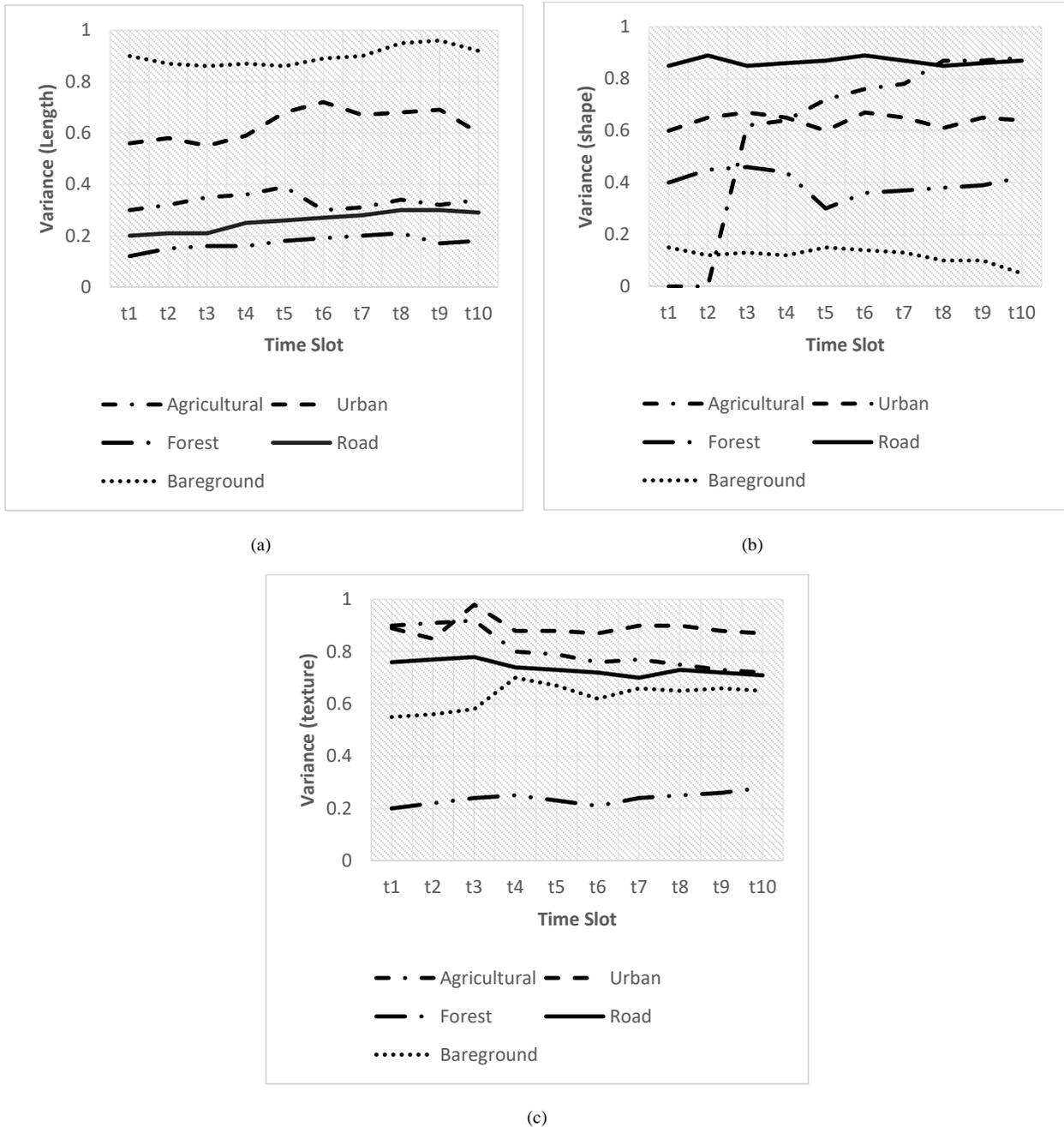

Fig 3. Variance of features associated with five objects in ST_Dataset_2 (a) length (b) shape (c) texture

The proportion of class changes over the period of time frame is given by Fig 4. In this graph, each bar presents the relative weighted amount of each class from the core tensor values as each time frame is increases. To see the overall gain and loss that has happened for the entire time period, Fig 5 provides the proper insight. The red colored portion indicates the percentage of loss that has happened to the particular class and the green colored portion indicates the percentage of gain that has happened to the particular class in the entire time frame under consideration.

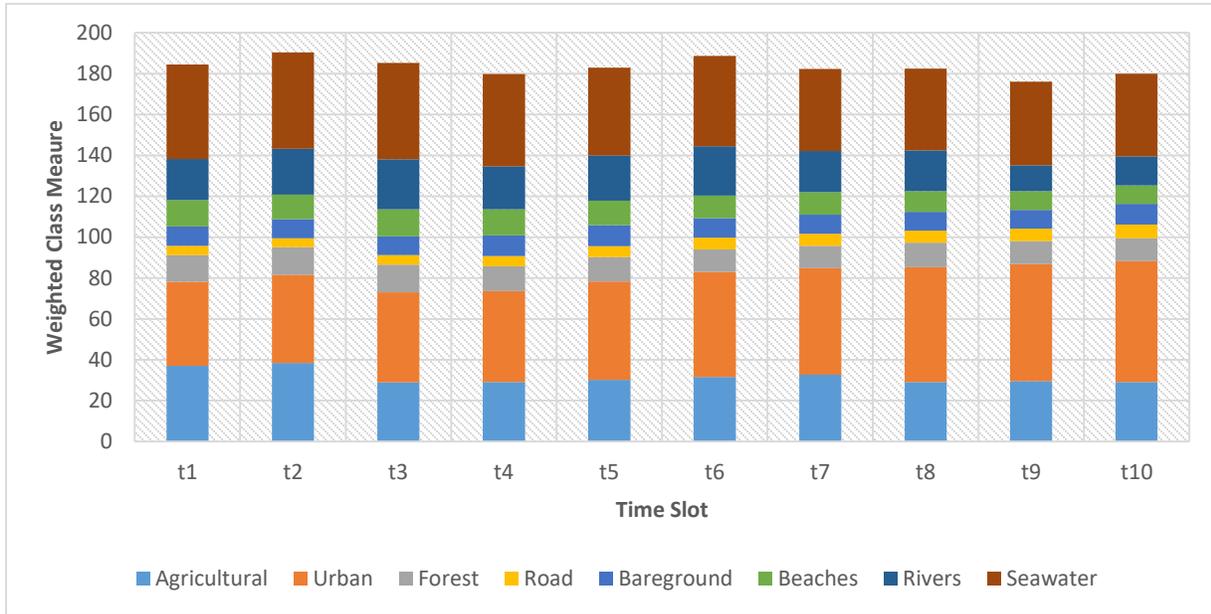

Fig 4. Proportion of class changes over the period of time frame for ST_Dataset_2

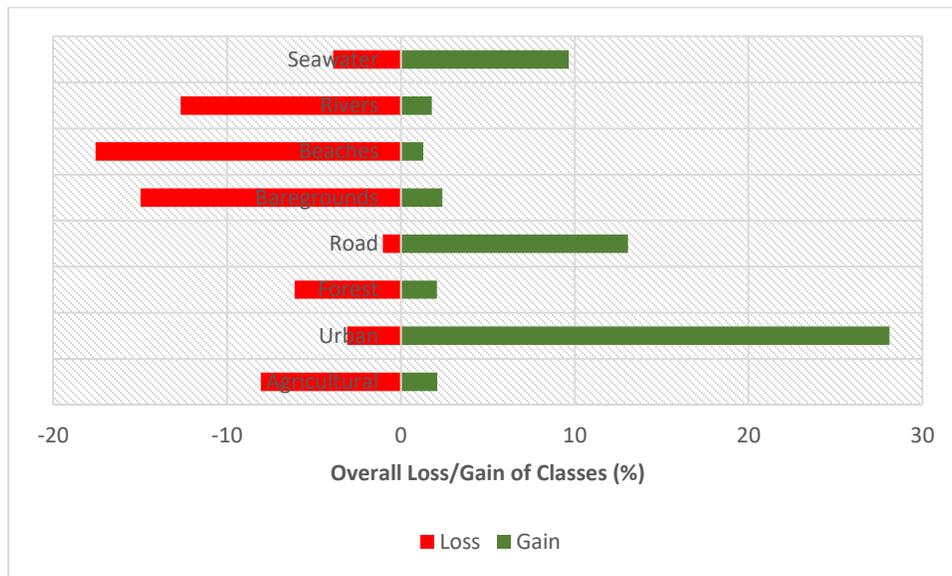

Fig 5. Overall loss/gain of classes over the period of time frame for ST_Dataset_2

## 5. DISCUSSIONS

This section attempts to make a comparison of the incremental Boolean Tensor Factorization with the traditional Boolean Tensor Factorization in terms of (a) Core tensor size (b) Factor Matrix Density and (c) Convergence Time.

Both the datasets ST_Dataset_1 and ST_Dataset_2 are evaluated for the core tensor size for both the approaches. The results are presented in Fig 6. It is seen that the reconstruction error is less for both the datasets in the same rank of core-tensor for the incremental approach. It can be deduced that the incremental approach converges faster as compared to the traditional method.

The convergence point of the datasets are also analyzed with respect to the factor matrix density. Fig 7 presents the plot of reconstruction error versus factor matrix density. Both the methods performs synonymously in these analysis for ST_Dataset_1 and ST_Dataset_2. The denser the data, the greater the reconstruction error, as expected. Data with dense values will obviously make the performance of algorithms more complex.

All the above said analysis points to the fact that the total convergence time for the incremental approach will be less as compared to the traditional approach. Fig 8 shows the experimental results wherein the convergence time for the Boolean Tensor Factorization for traditional and incremental approach is given for both the datasets. It is seen that the incremental approach converges faster when compared to the traditional approach. It is also noted that there is a steep rise in the change from first time slot to second time slot in incremental approach which owes to the computational cost of finding the variance of the factor matrices and associated diagonalization problem. However, after the second time slot, the convergence time does not increase drastically in the incremental approach.

The convergence time will also be dependent on the function of time used for updating the variance matrix. This work does not attempt to evaluate the results on the same, as the intention is to extent this work with different functions of time to model changes more semantically. There is a small rise observed in Incremental BTF when it moves from the first time slot to second time slot. This is due to the fact that the factor matrices and their variances are derived for the first instant and all the remaining computations are only updating the variance matrix.

All experiments were run on 3.3-GHz Centrino PC machine with 4G main memory. The algorithms were implemented in MATLAB using Matlab Tensor Toolbox.

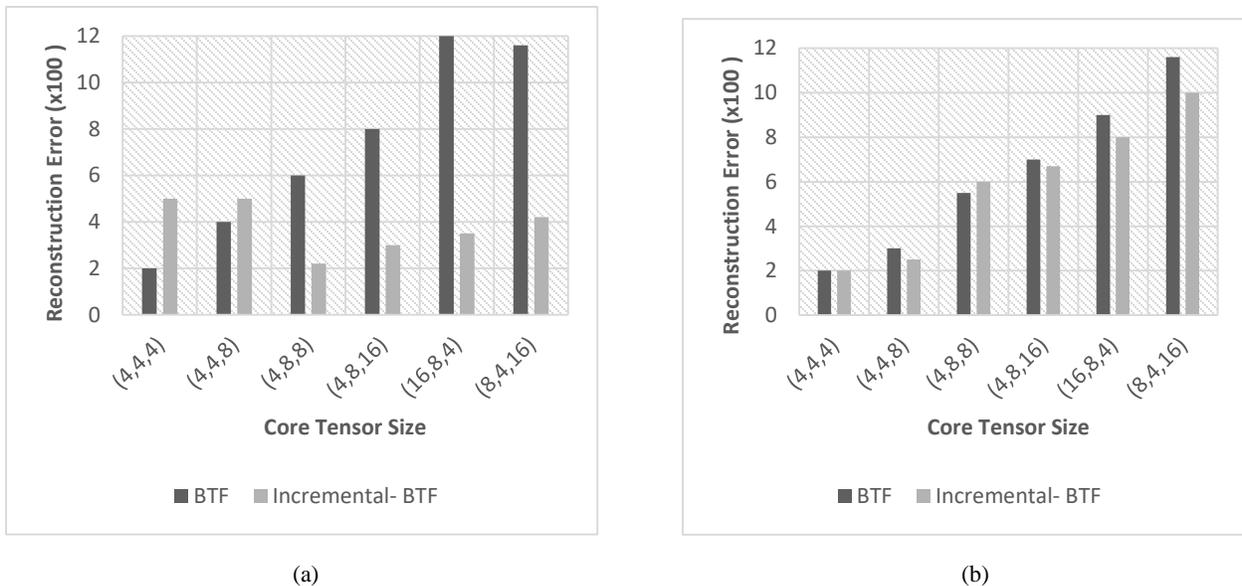

Fig 6. Reconstruction error for different core tensor size for (a) ST_Dataset_1 and (b) ST_Dataset_2

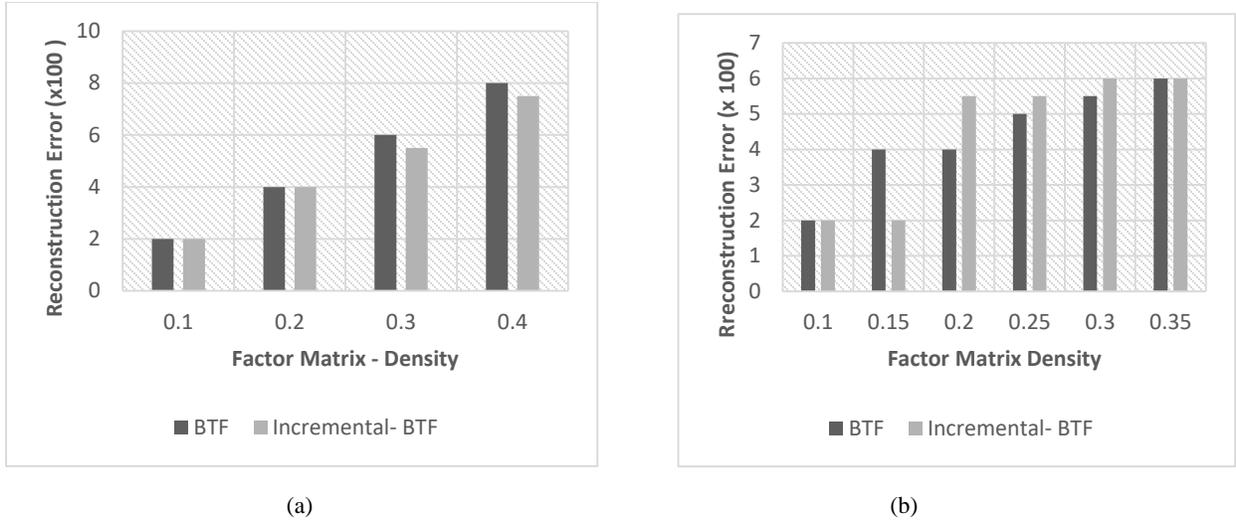

(a)                                          (b)

Fig 7. Reconstruction error for different factor density for (a) ST_Dataset_1 and (b) ST_Dataset_2

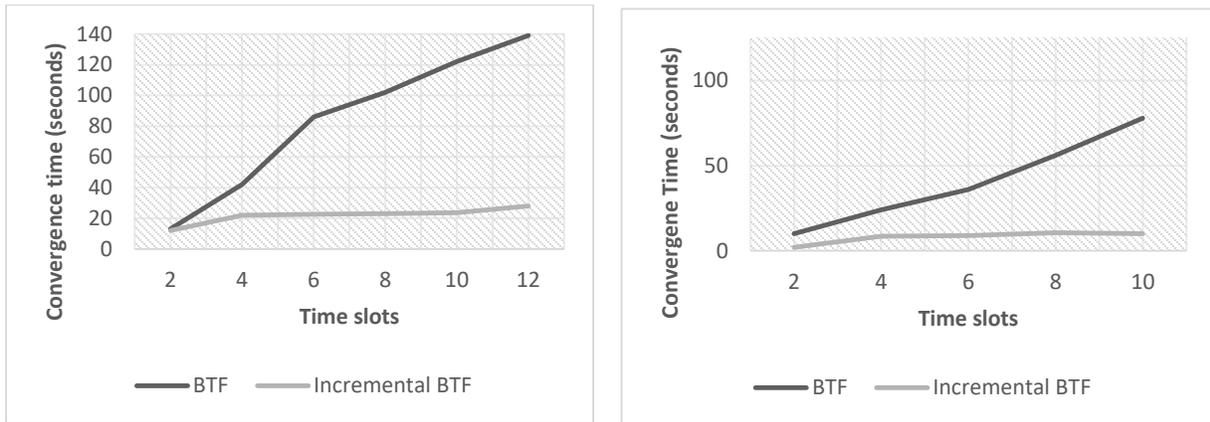

Fig 8. Convergence Time for (a) ST_Dataset_1 and (b) ST_Dataset_2

## 6. CONCLUSION

This paper proposes the modeling of change patterns in a spatiotemporal domain using an incremental Boolean Tensor Factorization approach. The computing speed of the algorithm is attributed to the fact that it deviates from traditional Boolean Tensor Factorization approach and adopts an incremental approach in finding the core-tensor and factor matrices. The execution time of the algorithm is highly dependent on the convergence of tensor factorization, which was seen to be less for the incremental approach as compared with the traditional approach. The change patterns modeled in this approach not only detect changes in objects in a whole, but also depicts the changes in the features of the objects. Obviously, the features can be selected as per the interest of the domain. The core tensor yielded after factorization will also show the overall changes which has happened to the objects in the area under consideration. There is also a notable save in the space for incremental approach as 'T' tensors associated with the time slots are not being stored. Further experiments and evaluations have to be oriented in the following directions (i) modeling change patterns using ontologies to bridge the semantic gap and (ii) observe the change patterns on different functions of time like seasonal, cyclic, and irregular by accumulating large dataset.